\documentclass[lettersize,journal]{IEEEtran}
\usepackage{amsmath,amsfonts}
\usepackage{algorithm}
\usepackage{array}
\usepackage[caption=false,font=normalsize,labelfont=sf,textfont=sf]{subfig}
\usepackage{textcomp}
\usepackage{stfloats}
\usepackage{url}
\usepackage{verbatim}
\usepackage{graphicx}
\usepackage{cite}
\usepackage{multirow}
\usepackage{graphicx}
\usepackage[table]{xcolor}
\usepackage{makecell}
\usepackage{float}
\usepackage[hidelinks]{hyperref}
\usepackage{algpseudocode}
\usepackage{caption}
\usepackage{color,soul}

\begin{document}

\title{DCNFIS: Deep Convolutional Neuro-Fuzzy Inference System}

\author{Mojtaba Yeganejou, Kimia Honari, Ryan Kluzinski,\\ Scott Dick,~\IEEEmembership{Member,~IEEE,} Michael Lipsett,~\IEEEmembership{Member,~IEEE,}, James Miller,~\IEEEmembership{Member,~IEEE}
	
\thanks{M. Yeganejou, K. Honari, R. Kluzinski, S. Dick, J. Miller are with the Dept. of Electrical \& Computer Engineering, M. Lipsett the Dept. of Mechanical Engineering, University of Alberta, Edmonton, AB, Canada, T6G 1H9. Emails: yegaenjo@ualberta.ca, honari@ualberta.ca, rkluzins@ualberta.ca, sdick@ualberta.ca, mike.lipsett@ualberta.ca, jimm@ualberta.ca. }
\thanks{This research was funded in part by the Natural Science and Engineering Research Council of Canada under grant nos. RGPIN 2017-05335 and CRDPJ 543705}}

% The paper headers
\markboth{MANUSCRIPT FOR IEEE TRANSACTIONS ON FUZZY SYSTEMS}%
{Yeganejou \MakeLowercase{\textit{et al.}}: DCNFIS: Deep Convolutional Neuro-Fuzzy Inference System}

\IEEEpubid{0000--0000/00\$00.00~\copyright~2024 IEEE}
% Remember, if you use this you must call \IEEEpubidadjcol in the second
% column for its text to clear the IEEEpubid mark.

\maketitle

\begin{abstract}
A key challenge in eXplainable Artificial Intelligence is the well-known trade-off between the transparency of an algorithm (i.e., how easily a human can directly understand the algorithm, as opposed to receiving a post-hoc explanation), and its accuracy. We report on the design of a new deep network that achieves improved transparency without sacrificing accuracy. We design a deep convolutional neuro-fuzzy inference system by hybridizing fuzzy logic and deep learning models, and show that it performs at least as accurately \textcolor{black}{as four existing  convolutional neural networks on five benchmark datasets, including the large-scale ILSVRC dataset}. \textcolor{black}{We furthermore show that our architecture outperforms state-of-the-art deep fuzzy systems}. We then exploit the transparency of fuzzy logic by deriving explanations, in the form of saliency maps, from the fuzzy rules generated by our architecture. We investigate the properties of these explanations in greater depth using the Fashion-MNIST dataset. 
\end{abstract}

\begin{IEEEkeywords}
Deep fuzzy systems, Explainable artificial intelligence, Deep learning, Machine learning, Fuzzy logic, Neuro-fuzzy systems.
\end{IEEEkeywords}

{
\addtolength{\tabcolsep}{5pt}
\renewcommand{\arraystretch}{1.3}

\begin{table}[H]
    \caption*{\textcolor{black}{Table of Symbols}}
    % \caption{Table of Symbols\label{symbols_table}}
    \centering
    {\color{black}\begin{tabular}{|c|c|}
        \hline
        $M_{ji}$ & Membership for fuzzy rule $i$ and input $j$ \\ \hline
        $\beta_{ji}$ and $\mu_{ji}$ & Parameters of Gaussian membership function\\ \hline
        $\omega_i$  & Firing strength of rule $i$ \\ \hline
        $\zeta_i$ & Logit value for class $i$ \\ \hline
        $y_i$ & Probabilities after applying softmax on $zeta$ \\ \hline
        $W_{ji}$ and $b_i$ &  parameters of antecedent functions \\ \hline
    \end{tabular}}
\end{table}
}

\section{Introduction}
\IEEEPARstart{D}{eep} Neural Networks (DNNs) are the heart of the “new” Artificial Intelligence (AI) (as exemplified by AlphaGo’s defeat of a human champion). DNNs are currently the most accurate solutions for many problems, including image recognition \cite{ref1}, Natural Language Processing (NLP) \cite{ref2}, speech recognition \cite{ref3}, robotic surgery \cite{surgical}, medical diagnosis \cite{retinapaper} and many others. However, knowledge in a DNN is encoded as a distributed pattern of potentially hundreds of billions of connection weights \cite{ref4}. This is an incomprehensible representation for human beings. Add to this the fact that some of their decisions seem counter-intuitive or even inexplicable to human experts, and it seems likely that humans would be reluctant to trust a DNN. Indeed, the literature bears this out \cite{ref5,ref6,ref7}. There is a risk that, far from embracing the AI revolution, human society might reject it if new mechanisms to foster trust in AI are not developed.

A classical approach to the interpretability problem is to incorporate an “explanation mechanism” in AI algorithms \cite{ref8,ref5,ref6,ref7}; such approaches have recently been dubbed eXplainable Artificial Intelligence (XAI). Explanations in turn enable knowledge discovery \cite{ref9}, algorithm verifiability \cite{ref10}, and even legal compliance (by satisfying the "right to an explanation" that increasingly appears in privacy protection legislation) \cite{ref11}. A discussion of the history of XAI techniques can be found in \cite{ref12} and some of the challenges of XAI are discussed in \cite{ref13,ref14,ref15}. Prior research in XAI for shallow learning focused on rule extraction from trained neural networks; see \cite{ref16} for a discussion. Recent work in this vein for deep networks includes rule extraction (e.g. \cite{ref17}), but seems to focus more upon visualizations such as saliency maps \cite{ref18,ref19}. Alternatively, one could directly design a deep network architecture to be more interpretable, following e.g. the ideas of fuzzy neural networks and neuro-fuzzy systems (e.g. the Adaptive Neuro-Fuzzy Inference System, ANFIS \cite{ref23}) \cite{ref20,ref21}. Interpretability is a key system goal for fuzzy systems and their hybrids \cite{ref22}, which in ANFIS is operationalized by mimicking the structure of a Takagi-Sugeno-Kang fuzzy inferential system. ANFIS can thus be directly translated to a fuzzy rulebase \cite{ref23}. However, there has long been a trade-off observed between algorithm interpretability and predictive accuracy; interpretable systems tend to be less accurate, while more accurate ones are less transparent, or even opaque to human understanding \cite{ref23,ref24}.

We thus seek to design a deep neural network that is simultaneously more interpretable than existing CNNs, while also remaining as accurate. Our solution (applicable to any CNN) is to first remove the final dense layers of the network, leaving only the network’s \emph{convolutional base} \cite{ref25}. We then concatenate a modified ANFIS neuro-fuzzy system \cite{ref21} to it as a new classifier (equivalently, we use the convolutional base as an automated feature extractor for the ANFIS classifier), producing a family of architectures we refer to as Deep Convolutional Neuro-Fuzzy Inferential Systems (DCNFIS) \cite{ref26}. \IEEEpubidadjcol With a few modifications to the ANFIS algorithm, we are able to perform end-to-end training on DCNFIS (unlike our previous deep fuzzy methods proposed in\cite{ref28,ref29,ref30});  \textcolor{black}{ our experiments used the ADAM optimizer \cite{ref27}, but any other optimizer can be used for the training of DCNFIS. A time complexity analysis indicates that the asymptotic growth for our classifier component is $O(N_C \cdot N_V)$, where $N_C$ is the number of classes in the dataset, and $N_V$ is the number of features at the end of the convolutional component}. Experiments on the MNIST Digits, Fashion-MNIST, CIFAR-10, and CIFAR-100 datasets using LeNet, ResNet, and Wide ResNet architectures indicate that DCNFIS models as accurate as the base CNNs they are built from. \textcolor{black}{Furthermore, when we compare the performance of DCNFIS with a wide range of recent deep and shallow fuzzy methods on the same datasets, we find that DCNFIS outperforms all of them. Thus, to the best of our knowledge, DCNFIS represents the state-of-the-art in deep neuro-fuzzy inference systems}.

\textcolor{black}{In another set of experiments, we evaluate DCNFIS on a large-scale image classification dataset. ImageNet Large Scale Visual Recognition Challenge (ILSVRC) \cite{russakovsky2015imagenet} is a commonly-used benchmark large-scale image classification dataset. For these experiments, we use a fourth CNN (Xception \cite{chollet2017xception}) as the backend of DCNFIS. To the best of our knowledge, these experiments are the first evaluation of a deep fuzzy system on this dataset. These experiments examine the scalability and accuracy of DCNFIS in large-scale learning. } 

\textcolor{black}{DCNFIS, as a deep neuro-fuzzy system, offers enhanced transparency as the DCNFIS classifier maps to a fuzzy rulebase. In this paper, we adapt the cluster-medoid approach we proposed in \cite{ref28,ref29,ref30} to the fuzzy rules induced by DCNFIS. We treat the fuzzy region defined by each rule as a cluster, and determine the medoid element of that cluster. We then generate a saliency map for that element, which we propose as an explanation for the whole cluster. We investigate the properties of these explanations using the Fashion-MNIST dataset as a case study.}

Our contributions in this paper are, firstly, the design and evaluation of a family of novel hybrid fuzzy deep-learning algorithms. Secondly, we design an explanation mechanism leveraging the properties of fuzzy rule-based systems, and explore its properties.

The remainder of this paper is organized as follows. In Section 2 we review essential background on deep learning and neuro-fuzzy systems, \textcolor{black}{and conduct a critical inquiry into the concept of "trusted AI," and how XAI contributes to it. In Section 3 we present our proposed architecture and our methodology for evaluating its performance. Our experimental results characterizing the accuracy of the method for small datasets and large-scale dataset are presented in Section 4 and 5 respectively. Our explanation mechanism is designed and its properties investigated in Section 6. We close with a summary and discussion of future work in Section 7.}

\section{Background Review}
\subsection{Fuzzy Classifiers}
Fuzzy classifiers \cite{ref32,ref33} assume the boundary between two neighboring classes is a continuous, overlapping area within which an object has partial membership in each class. These classifiers provide a simple and understandable representation of complex models using linguistic if-then rules. The rules are of the basic form: 
\begin{equation}
	\label{eq_1}
	if \ X_1  \ is \ A \ and \ X_2 \ is \ B \ then \  Z \ is \ C
\end{equation}

where $X_1$ and $X_2$ are the input variables of the classifier; $A$ and $B$ are linguistic terms \cite{ref34}, characterized by fuzzy sets \cite{ref35}, which describe the features of an object; and $C$ is a class label. The firing strength of this rule with respect to a given object represents the degree to which this object belongs to the class $C$.

\subsection{Convolutional Neural Networks}
Modern Convolutional Neural Networks (CNNs) were first described by LeCun et al. in \cite{ref36}. In that work, the network consisted of stacks of alternating convolution and pooling layers only. Empirically, feature extraction via deep CNNs is currently the leading approach for building accurate neural models \cite{ref36,ref37,ref38,ref39}. More recent CNN architectures have added additional operations or layers to incorporate additional properties. For example, Local Response Normalization (LRN) layers have been added to CNNs in order to implement the concept of lateral inhibition from human vision \cite{ref37}. One particular example is the ResNet architecture \cite{ref40}, which focuses on correcting an observed degradation in training accuracy for CNNs with a large number of layers \cite{ref1}. The ResNet approach is to change the transfer function of the layer, by adding shortcut links. The network is trained to mimic the mapping $F(X)=H(X)-X$, where $H(X)$ is the actual optimization target. The original input/output mapping is refactored as $F(X) + X$. Empirically, this residual appears easier to learn than the original mapping. An updated version of ResNet was recently published in \cite{ref41}. Wide Residual Networks (WRN)\cite{ref42} are another version of ResNets. ResNet architectures are very deep, and they have the problem of diminishing feature reuse, which makes these networks very slow. WRN architectures are wider (more feature maps per convolutional layer) while they have less depth (fewer convolutional layers), resulting in faster training.

\textcolor{black}{The Xception \cite{chollet2017xception} model is another recent deep convolutional neural network architecture based entirely on depthwise separable convolution layers. The model is built on the hypothesis that the mapping of cross-channels correlations and spatial correlations in the feature maps of convolutional neural networks can be entirely decoupled. This idea is a further development of the Inception \cite{szegedy2016rethinking}. The Xception architecture has 36 convolutional layers forming the feature extraction base of the network. These convolutional layers are structured into 14 modules with linear residual connections around them, except for the first and last modules.}   

\subsection{Deep learning and fuzzy systems}
Hybridizations of fuzzy logic and shallow neural networks have been investigated for over 25 years \cite{ref43}. Hybridizations of fuzzy logic and deep networks, on the other hand, have only received substantial attention in the last decade. As with the older literature, we can distinguish between fuzzy neural networks - in which the network architecture remains the same, but neuron-level operations are fuzzified in some way – and neuro-fuzzy systems wherein the network architecture is altered to mimic fuzzy inference algorithms \cite{ref43}. 

Fuzzifying inputs to a deep MLP network is done in \cite{ref44}. Chen et al.’s fuzzy restricted Boltzmann machine \cite{ref45} is refactored to use Pythagorean fuzzy sets \cite{ref46} by Zheng et al. \cite{ref47}, and interval type-2 fuzzy sets by Shukla et al. \cite{ref48}. They were also stacked to form a fuzzy deep belief network in \cite{ref49}. Zheng et al. hybridized Pythagorean fuzzy sets and stacked denoising autoencoders in \cite{ref50}. \textcolor{black}{ \cite{sharma2019fuzzy} and \cite{diamantis2020fuzzy} present new fuzzy pooling operations and investigate their performance in the context of image classification. \cite{beke2019interval} presents an activation function, called Interval Type-2 (IT2) Fuzzy Rectifying Unit (FRU) and employs it to improve the performance of the Deep Neural Networks. \cite{bodyanskiy2022deep} proposes a fuzzy neuron (F-neuron) as an adaptive alternative for the existing piece-
wise activation functions. In \cite{sharma2023mixed}, a mixed fuzzy pooling operation is proposed for image classification in the CNN architecture}. Echo state networks were modified to include a second reservoir in \cite{ref51}, which uses fuzzy clustering for feature reinforcement (combating the vanishing gradient problem). Layers of these fuzzy echo state networks are then stacked into a deep network. Stacked fuzzy rulebases, learned via the Wang-Mendel algorithm, are proposed in \cite{ref52}. An evolving fuzzy neural network for classifying data streams is presented in \cite{ref53}. A deep fuzzy network for software defect prediction, which incorporates stratification to rectify the class imbalance problem, is proposed in \cite{ref54}. The algorithm can add or merge layers in response to concept drifts within the stream. Fuzzy clustering is merged with the training of stacked autoencoders by adding constraints the optimize compactness and separation of clusters and within-class affinity to the network’s objective function in \cite{ref55}. A variation on this theme is using fuzzy logic as a preprocessor; in \cite{ref56}, pixels in an image were mapped to triangular fuzzy membership functions, which are three-parameter functions. Feature maps consisting of each of those parameters were then passed to three different NNs, and their outputs fused. Classification errors were reduced on several benchmark datasets. Fuzzy c-means clustering was used to preprocess images for stacked autoencoders in \cite{ref57}. Fuzzy logic is woven into a ResNet model for segmenting lips within human faces in \cite{ref58}. The Softmax classification layer is replaced with a fuzzy tree model trained using fuzzy rough set theory in \cite{ref59}. Aviles et al. \cite{ref60} engineered a hybrid of ANFIS, recurrent networks and the Long Short-Term Memory for the contact-force problem in remote surgery. Rajurkar and Verma \cite{ref61} design stacked TSK fuzzy systems, while \cite{ref62} is a more complex stacked TSK with a focus on interpretability. An adversarial training algorithm for a stacked TSK system is proposed in \cite{ref63}. Tan et al. suggest using fuzzy compression to prune redundant parameters from a CNN \cite{ref64}. A specialized neural network for solving polynomial equations with Z-number coefficients is proposed in \cite{ref65}. A neural network that mimics the fuzzy Choquet integral is proposed in \cite{ref66}. John et al. \cite{ref67} employ FCM-based image segmentation to preprocess images before training a CNN, while \cite{ref68} and \cite{ref69} instead used the deep network (restricted Boltzmann machines and Resnet, respectively) as a feature extractor prior to clustering. This last is quite similar to the approach in \cite{ref30} and \cite{ref29}: the densely-connected layers at the end of a deep network are replaced with an alternative classifier built from a clustering algorithm. The latter two, however, employ the fuzzy C-means and Gustafson-Kessel fuzzy clustering algorithms, respectively. \textcolor{black}{ \cite{riaz2019semi} introduces a semi-supervised approach to fuse fuzzy-rough C-mean clustering with CNNs. Yazdanbakhsh et al. \cite{yazdanbakhsh2019deep} introduced the first end-to-end deep neuro-fuzzy structure for image classification by developing two new operations: fuzzy inference operation and fuzzy pooling operations. A preliminary report on DCNFIS was published in \cite{ref26}. \cite{Zhang2018} proposed a highly interpretable deep TSK fuzzy classifier HID-TSK-FC based on the concept of shared linguistic fuzzy rules. \cite{zhou2017stacked} designs a new stacked-structure-based hierarchical TSK fuzzy classifier and calls it SHFA-TSK-FC claiming to have promising performance and high interpretability at the same time. They furthermore investigate the shortcoming of the existing hierarchical fuzzy classifiers in interpreting the outputs and fuzzy rules of intermediate layers. \cite{gu2018semi} introduces a semi-supervised learning approach using a deep rule-based classifier that generates human interpretable if-then rules. \cite{tan2023deep} develops an evolutionary unsupervised learning representation model with iterative optimization by implementing a deep adaptive fuzzy clustering method, which learns the behaviour of a convolutional neural network classifier from given only unlabeled data samples. \cite{guo2021concise} presents an enhanced ANFIS which integrates improved bagging and dropout to build concise fuzzy rule sets. \cite{shah2020adaptive} proposes an adaptive fuzzy network based convolutional network which uses a pre-trained deep convolutional network with a subsequent adaptive fuzzy based network.  In \cite{di2021advanced}, a fuzzy relational neural network based model is presented which extrapolates relevant information from images to obtain a clearer indication on the classification processes. \cite{xi2018interpretable} focuses on interpretability of machine learning and adds Neural-Fuzzy based interpretability-oriented layers to the CNNs in the form of Fuzzy Logic-based rules. \cite{zhang2019deep} presents a deep fuzzy k-means (DFKM) with adaptive loss function and entropy regularization. }

\subsection{\textcolor{black}{Trust and XAI}}
\textcolor{black}{If the goal of XAI is to foster user trust in AI algorithms, then we seem far from achieving it. Findings in \cite{ref70} indicate that AI is only entrusted with low-risk, repetitive work \cite{ref70,ref71}. As one study participant stated, “I don’t think that an AI, no matter how good the data input into it, can make a moral decision.”\cite{ref71} However, “trust in AI” is not a simple thing. Generally, trust is a complex, mercurial, situation-dependent concept that defies easy definition. A review in \cite{ref72} found that trust was defined variously as a social structure, a verb, a noun, a belief, a personality trait, etc. It has been studied in the fields of psychology, sociology, organizational science, and information technology (IT), among others \cite{ref73}. One common thread across these domains is that the act of extending trust necessarily renders the person extending it (the trustor) vulnerable to the person being trusted (the trustee) \cite{ref74,ref75,ref76}. Extending trust is thus not a trivial act, and a number of factors influence a person’s decision to trust another or not. }

\textcolor{black}{What, then, are the implications for XAI? Let us first consider how trust is defined. Several definitions appear in the XAI literature. There is of course the classic meaning of trust as a willingness to be vulnerable to another entity \cite{ref77}. In a definition more directly aimed at autonomous systems, Lee and See define trust as an “attitude that an agent will help achieve an individual’s goals in a situation characterized by uncertainty and vulnerability” \cite{ref71,ref82}; this seems to be the most commonly adopted definition in the XAI literature.  Israelsen and Ahmed define it as “a psychological state in which an agent willingly and securely becomes vulnerable, or depends on, a trustee (e.g., another person, institution, or an autonomous intelligent agent), having taken into consideration the characteristics (e.g., benevolence, integrity, competence) of the trustee” \cite{ref83}. Others take a more behaviorist slant; trust was “the extent to which a user is confident in, and willing to act on the basis of, the recommendations, actions, and decisions of an artificially intelligent decision aid” in \cite{ref84}.}

\textcolor{black}{In this view, XAI - and explanations in particular - are a means of persuading users to be vulnerable to an AI, in exchange for the AI’s services. As Miller points out, explanation is a social act; an explainer communicates knowledge about an explanandum (the object of the explanation) to the explainee; this knowledge is the explanans \cite{ref85}. In XAI, a decision or result obtained from an AI model is the explanandum, the explainee is the user who needs more information, the explainer is the Explanation Interface (EI) of the AI, and the explanans is the information communicated to the user about the decision / result \cite{ref86}.}

\textcolor{black}{These definitions, however, leave out the basic question of “whom are we trusting?” The Locus of Trust (LoT, the person being trusted), is key to the trusting relationship. A website, for instance, is created and completely controlled by humans. They are clearly the LoT \cite{ref91,ref92,ref93}. A large deep network, on the other hand,  has almost certainly discovered and exploited patterns in its input data that were unknown to the human analyst. The AI's designers have influence over it, but typically do not fully understand the induced model. So is the LoT the AI itself? Some researchers assert this, e.g. \cite{ref94,ref78,ref83,ref84}. Some evidence also indicates that users themselves react to AI agents as persons, e.g. \cite{ref91,ref96}.}

\textcolor{black}{Alternatively, we can view the LoT as being the designers of an AI. It is the designers, after all, who determine the data (and implicitly the biases) used in the training of a machine learner. They determine how fairness, ethics, transparency, and safety are realized within the AI. They will monitor and correct the learning process for the AI, and its ongoing adaptation in real-world usage. Thus, \cite{ref97} and others argue that the human designers should be the LoT, and held responsible for the AI’s behavior. Furthermore,  accountability for the AI's decisions is an aspect of trustworthiness \cite{ref98,ref99}; one that must be borne by humans.}

\section{Methodology}

\subsection{Proposed Architecture}
\begin{figure}[!t]
	\centering
	\vspace*{-5mm}\hspace*{-6mm}\includegraphics[width=\linewidth,trim={45 30 70 40},clip]{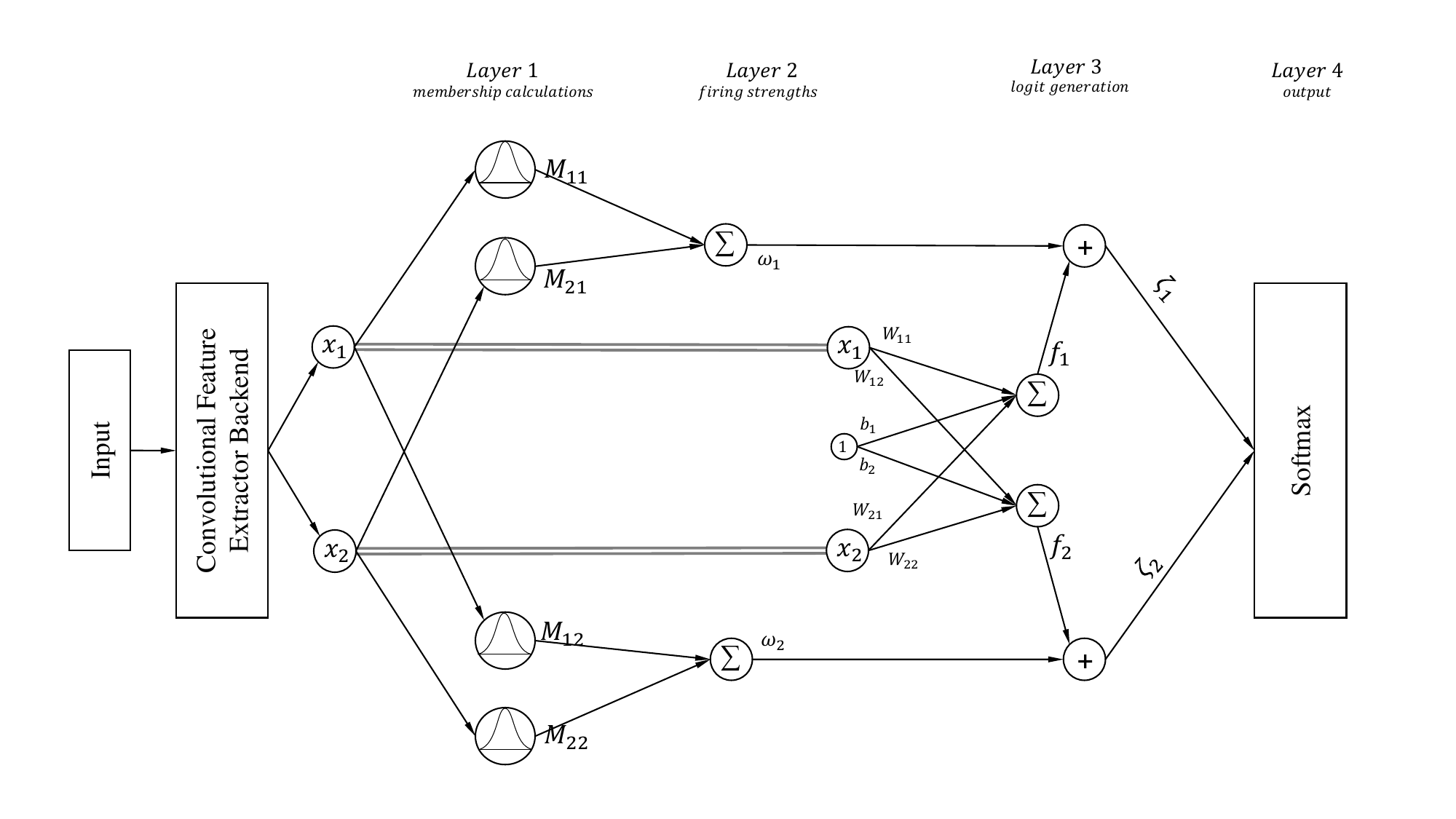}
	\caption{\textcolor{black}{Simplified Block Diagram of our Deep Convolutional Neuro-Fuzzy Classifier}}
	\label{fig_1}
\end{figure}

Our fuzzy classifier is based on the Adaptive Neuro-Fuzzy Inference System (ANFIS) architecture \cite{ref21,ref103}. ANFIS is known to be a universal approximator \cite{ref21}, and is thus in theory capable of reproducing the mapping realized by the fully-connected layer(s) it replaces in the base CNN architectures. ANFIS is a layered architecture where the first layer computes the membership of an input in a fuzzy set (in this case, defined by a Gaussian membership function) using Eq. \ref{eq_2}, for fuzzy rule $i$ and input $j$. There is one membership per input for each rule; Fig. \ref{fig_1} shows a system with 2 inputs and 2 fuzzy rules, resulting in 4 membership functions. We compute the natural logarithm of the membership values, for numerical stability in the tails. The following equations compare our exact implementation with ANFIS equations.
\textcolor{black}{
\begin{equation}
	\label{eq_2}
	\begin{array}[hbt]{c}
		\displaystyle M_{ji} = \exp(-\beta_{ji} \cdot (x_j - \mu_{ji})^2) \ \ \ \ \ \ \ \ \ _{(ANFIS)}\\
		\displaystyle M_{ji} = -0.5 \cdot \beta_{ji} \cdot (x_j - \mu_{ji})^2 \ \ \ \ \ \ \ \ \  _{(DCNFIS)}
	\end{array}
\end{equation}
}

Layer 2 computes the firing strength of each fuzzy rule. This is usually the product of all antecedent membership functions; however, as we are using logarithms, we sum the log memberships
\textcolor{black}{
\begin{equation}
	\label{eq_3}
	\begin{array}[hbt]{c}
		\displaystyle \omega_i = \prod_{j} M_{ji} \ \ \ \  _{(ANFIS)}\\
		\displaystyle \omega_i = \sum_{j} M_{ji} \ \ \ \ _{(DCNFIS)}
	\end{array}
\end{equation}
}
In layer 3, the activation is normally computed as a linear combination of the input variables, multiplied by the firing strength of the respective rule. This is changed to the sum of the log firing strength and the linear combination of the input variables. \textcolor{black}{The outputs of layer 3 are called logit values.}

\textcolor{black}{
\begin{equation}
	\label{eq_4}
	\begin{array}{c}
		\displaystyle \zeta_i = \omega_i \cdot (\sum_j W_{ji} \cdot x_j + b_i ) \ \ \ \ \ _{(ANFIS)} \\
		\displaystyle \zeta_i = \omega_i +(\sum_j W_{ji} \cdot x_j + b_i ) \ \ \ 	_{(DCNFIS)}
	\end{array}
\end{equation}
}
\textcolor{black}{Note that we do not take the logarithm of the linear combination of inputs. This modification allows us to create fuzzy regions (clusters) along with an end-to-end trainable deep convolutional backend without causing an error gradient explosion.}

Finally, layer 4 computes the class probabilities using the softmax activation function.
\textcolor{black}{
\begin{equation}
	\label{eq_5}
	\begin{array}{c}
		\displaystyle P\left(c_i\middle|\ x\right)=y_i=\frac{\exp{\left(\zeta_i\right)}}{\sum_{o}{\exp{\left(\zeta_o\right)}}}
	\end{array}
\end{equation}
}

This architecture mimics the operation of a Fuzzy Inferential System (FIS), which is a rule-based expert system (and thus highly interpretable). Specifically, the layer transfer functions mimic the stages of processing by which an FIS infers an output from its inputs. The hybrid learning rule for ANFIS (a Kalman filter in Layer 4, gradient descent for Layer 1) allows the fuzzy rule-base to be induced from a dataset. There is a 1:1 correspondence between ANFIS and the FIS it mimics, and so we can directly translate a trained ANFIS into fuzzy rules. However, this only applies to the direct inputs of ANFIS - which are the outputs of the final CNN layer. 

\begin{algorithm}[ht]
	\caption{\textcolor{black}{Modified ANFIS}}\label{alg1}
	{\color{black}\begin{algorithmic}
		\Require \\
		{$N_C\gets$ Number of output classes (number of rules)}\\  
		{$N_V\gets$ Number of extracted features by CNN (number of input variables to ANFIS)}\\
		{$M_{ji} \gets$ Membership value of variable $j$ to membership functions of rule $i$}
		\State
		\State{// \textbf{Layer 1 and 2}}
		\For {$i$ from 1 to $N_C$}
		\State $\omega_i=0$
		\For {for $j$ from 1 to $N_V$}
		\State $M_{ji}= -0.5 \cdot \beta_{ji} \cdot (x_j  -\mu_{ji})^2$  
		\State $\omega_i=\omega_i+M_{ji}$
		\EndFor   
		\EndFor\\
		% \State{// \textbf{Layer 2}}\\
		% $\Bar{\omega} = 0$
		% \For {$i$ from 1 to $N_C$}
		% \State $\Bar{\omega} = \Bar{\omega} + \omega_i$
		% \EndFor
		% \For {$i$ from 1 to $N_C$}
		% \State $\Bar{\omega_i} = \frac{\omega_i}{\Bar{\omega}}$
		% \EndFor
		
		\State{// \textbf{Layer 3}}
		
		\For {$i$ from 1 to $N_C$}
		\State $f_i = 0$
		\For {for $j$ from 1 to $N_V$}
		\State $f_i = f_i + W_{ji} \cdot x_j$
		\EndFor
        \State $f_i = f_i + b_i$
		\EndFor\\
		\For {$i$ from 1 to $N_C$}
		\State $\zeta_i = \omega_i + f_i$
		\EndFor\\
		\State{// \textbf{Layer 4}}\\
		\State Send $\zeta_i$ to softmax activation function accourding to e.q. \ref{eq_5}
		
	\end{algorithmic}}
\end{algorithm}

\begin{algorithm}[!h]
	\caption{\textcolor{black}{Medoid extraction from fuzzy clusters}}\label{alg2}
	{\color{black}\begin{algorithmic}                
		\State $rules\gets$ inference on $model.layers['rules']$
		\State $features\gets$  inference on $model.layers['features']$
		% \State $cluster\_labels \gets$ harden the fuzzy clusters out of $rules$ 
		\For{each sample in $x\_train$}
		\State $cluster\_labels \gets$ maximum $rules$ (hardening)
		\EndFor
		\For{samples in each cluster}
		\State find distance of each sample to the rest
		\State sum up the distances
		\State $cluster\_medoid \gets$ the sample with lowest distance
		\EndFor
	\end{algorithmic}}
\end{algorithm}

Note that ANFIS does not provide a shortcut to the softmax layer. Layer 3 of ANFIS implements a linear combination of the network inputs for each individual rule. In the basic ANFIS, the rule is weighted by its firing strength (computed in Layer 2), and then passed to a summation. In our implementation, we have taken the logarithm of the network signals, and so we add the log of the firing strength to consequent functions in Layer 3, and then pass the logit value to the softmax function in Layer 4. \textcolor{black}{As the logarithm function is monotonic increasing, the ranks of the softmax outputs should be the same as for the original ANFIS, even if the logit values have changed. Algorithm \ref{alg1} shows the feed forward calculations for DCNFIS in detail. Based on this algorithm order of growth for the classifier of DCNFIS is $O(N_C \cdot N_V)$. Our FLOPS comparison using model-profiler \footnote{https://pypi.org/project/model-profiler/} shows no difference between DCNFIS ResNet29\_V2 and it's regular version which is used for Fashion-MNIST dataset. However, we would in general expect that DCNFIS will have more parameters (for calculating membership functions) in comparison with the regular CNN if the CNN's classifier is not a multi-layer perceptron with hidden layers. In these situations, the number of extra parameters of DCNFIS can be calculated by $2*N_C*N_V$ which relates to the $\beta$ and $\mu$ parameters of membership functions for the rule-generation part of the classifier.}

\subsection{Learning Algorithm}
Our fuzzy classifier will use the Adam \cite{ref27} algorithm for learning the adaptive parameters of the network. Adam optimization is a stochastic gradient descent method that is based on adaptive estimation of first-order and second-order moments. Thus, in order to update the parameters ($\theta$) ADAM first calcualtes gradients at time-step $t$, and then based on values of $\beta_1$ and $\beta_2$ bias-corrected first $(\hat{m_t})$ and second $(\hat{v_t})$  moment parameters will be updated. Finally the parameters will be updated based on: $\theta_t = \theta_{t-1} - \alpha \cdot \frac{\hat{m_t}}{\sqrt{\hat{v_t}+\epsilon}}$ where $\alpha$ is the step-size and $\epsilon = 10e-8$. The gradients passed to the convolutional layers are calculated as follows:

\begin{equation}
	\label{eq_6}
	\begin{array}{c}
		\displaystyle \frac{\partial\zeta_i}{\partial x_j} = \frac{\partial}{\partial x_j}{\omega_i}+\sum_{j}\ W_{ji}\ \\
		\textcolor{black}{\displaystyle  =\frac{\partial}{\partial x_j}{\left(M_{ji}\right)}+\sum_{j}\ W_{ji}} \\ \textcolor{black}{\displaystyle  =-\beta_{ji} \cdot \left(x_j-\mu_{ji}\right)+\sum_{j}\ W_{ji}}
	\end{array}
\end{equation}

Where the gradients of the membership function parameters can be calculated as follows:

\begin{equation}
	\label{eq_7}
	\begin{array}{cccc}
		\vspace{1mm} \displaystyle \frac{\partial\zeta_i}{\partial \omega_i}=1 \\
		\vspace{1mm} \displaystyle \frac{\partial\zeta_i}{\partial  M_{ji}}=1 \\
		\textcolor{black}{\vspace{1mm} \displaystyle \frac{\partial\zeta_i}{\partial\mu_{ji}}=\beta_{ji}\left(x_j-\mu_{ji}\right)} \\
		\textcolor{black}{\vspace{1mm} \displaystyle \frac{\partial\zeta_i}{\partial\beta_{ji}}=-0.5 \cdot \left(x_j-\mu_{ji}\right)^2}
	\end{array}
\end{equation}

\subsection{\textcolor{black}{Guided Backpropagation}}
\begin{figure}[!t]
	\centering
	\hspace*{-6mm}\includegraphics[width=0.6\linewidth,trim={75 150 595 60},clip]{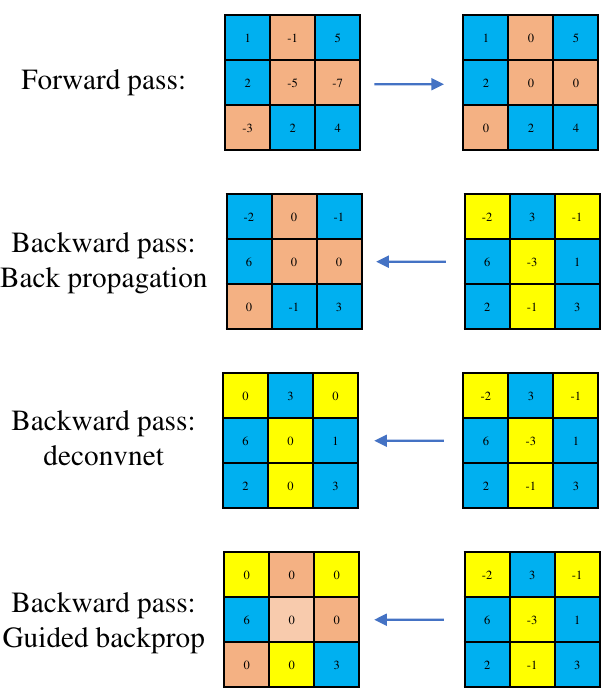}
	\caption{\textcolor{black}{A simple visualization on guided backpropagation}}
	\label{fig_guide}
\end{figure}

\textcolor{black}{Guided backpropagation \cite{ref104}, is a variant of the “deconvolution approach” for visualizing features learned by CNNs. Guided backpropagation allows the flow of only the positive gradients by changing the negative gradient values to zero. Fig. \ref{fig_guide} shows a simple visualization on how guided backpropagation works. }

\textcolor{black}{In order to choose the saliency analysis method, we have measured RISE metrics \cite{petsiuk2018rise}, as implemented in Xplique package \cite{fel2022xplique}. Guided back-propagation had the highest score, and so is chosen for visualizing the medoids of our rule clusters determined in Algorithm 2. Our visual analysis also showed that guided back-propagation was the best choice. For details about comparing saliency analysis methods please see \cite{ref31,ref12,petsiuk2018rise}.} 

\section{\textcolor{black}{Evaluation on Benchmark Datasets}}

\subsection{Datasets}
We have chosen four commonly-used benchmark datasets for deep learning: MNIST \cite{ref100}, Fashion MNIST \cite{ref101}, CIFAR-10 \cite{ref37}, and CIFAR-100 \cite{ref102}. MNIST is a collection of grayscale images, each representing one handwritten digit and labeled with the correct value. The samples were gathered from roughly 250 individuals; one half are U.S. Census Bureau employees, while the other half are high school students. MNIST is pre-partitioned into 60,000 training samples and 10,000 testing samples; the writers for the two groups are disjoint. Each MNIST image is a 28x28 pixel square, centered on the center of mass of the digit pixels. These images are constructed from a set of 20x20 pixel square images containing the actual digits, size-normalized to this bounding box. While the original samples were binary images, the anti-aliasing technique used in the normalization introduced gray levels. See \cite{ref100} for more details. Fashion MNIST shares the same image size and structure of training and testing splits with MNIST. Each Fashion MNIST sample is a 28x28 grayscale image, associated with a label from one of ten classes (t-shirts/tops, trousers, pullovers, dresses, coats, sandals, shirts, sneakers, bags, and ankle boots) \cite{ref101}. CIFAR-10 \cite{ref37} is an RGB image dataset, consisting of 60,000 images of size 32x32 in ten mutually exclusive classes, with 6,000 images per class, again pre-partitioned into training and testing sets. There are 50,000 training images and 10,000 testing images. CIFAR-100 \cite{ref102} is similar to CIFAR-10, except it has 100 classes. This dataset is also pre-partitioned into 50,000 training images and 10,000 testing images.

\subsection{Experimental Setup}
The existing literature on MNIST, Fashion-MNIST, CIFAR-10, CIFAR-100 datasets reports results based on a single-split design; for comparability we will follow this design, employing the same test set of 10,000 images as our out-of-sample evaluation. We test LeNet \cite{ref36} with our fuzzy classifier, then ResNet and Wide ResNet. Our tests employ the second version of ResNet20 \cite{ref41} for MNIST, ResNet29 for Fashion MNIST, and ResNet164 for CIFAR-10 and CIFAR-100. We have tested WRN-28-10 for CIFAR-10, and WRN-28-12 for CIFAR-100 \cite{ref42}.

The LeNet-based architectures, are trained for 200 epochs in batches of 128. The parameters for the Adam optimizer have been set as $\eta = 0.001$, $\beta_1 = 0.9$, $\beta_2 = 0.999$. For the ResNet and WRN-based experiments the models were trained for 200 epochs in batches of 128. The learning rate has been set to 0.001 initially, and then it will be reduced to 1e-6 after epoch 180. Data augmentation techniques are applied on all datasets except MNIST. All the experiments are replicated ten times.

\subsection{Evaluation of Results}
As shown in TABLE \ref{table1}, the performance difference between the original CNNs and their DCNFIS versions is very minor; about half the time, the base architecture was slightly better, and half the time the DCNFIS was better. This is a significant change from \cite{ref28} and \cite{ref29}, in which the enhanced interpretability of the deep fuzzy system came at the price of clearly losing some overall accuracy.

We next conduct a statistical analysis of these results. As there are only ten replicates of each experiment, but the variances for the original CNN and the DCNFIS versions can be substantially different, we employ the $t$-test for unequal variances. At a significance level of $\alpha$ = 0.05, there are only two instances where the differences were significant: CIFAR-10 with ResNet, and CIFAR-100 with WRN. We thus claim that \textit{using the DCNFIS method does not appear to reduce the accuracy of a CNN.}
{
\setlength\arrayrulewidth{0.25mm}
\setlength\extrarowheight{3pt}
\begin{table}[!t]
	\caption{DCNFIS and Regular-CNNs Accuracy Comparison \label{table1}}
	\centering
	\begin{tabular}{ccccc}
		\hline 
		\multirow{3}{*}{Dataset}  & \multicolumn{4}{c}{LeNet}\\
		\cline{2-5} 
		& \multicolumn{2}{c}{Regular} & \multicolumn{2}{c}{DCNFIS}\\		
		\cline{2-5}
		& AVG & STD & AVG & STD\\
		\hline
		MNIST & 99.21 & 0.0404 & \textbf{99.23} & 0.462 \\
		\hline
		 Fashion-MNIST & \textbf{90.36} & 0.2364 & 90.10 & 0.1445 \\
		\hline
		CIFAR-10 & 73.11 & 0.4550 & \textbf{73.39} & 0.9267\\
		\hline
		CIFAR-100 & 43.25 & 0.6594 & \textbf{43.30} & 0.5472\\
		\hline
		& \multicolumn{4}{c}{ResNet}\\\cline{2-5}
		& \multicolumn{2}{c}{Regular} & \multicolumn{2}{c}{DCNFIS}\\\cline{2-5}
		& AVG & STD & AVG & STD\\
		\hline
		MNIST & 99.57 & 0.381 & \textbf{99.59} & 0.0288 \\
		\hline
		Fashion-MNIST & \textbf{94.64} & 0.0816 & 94.4 & 0.2308\\
		\hline
		CIFAR-10 & \textbf{93.13}$^a$ & 0.0844 & 93.02 & 0.0526\\
		\hline
		CIFAR-100 & 74.52 & 0.0978 & 74.52 & 0.1589\\
		\hline		
		& \multicolumn{4}{c}{WRN}\\\cline{2-5}
		& \multicolumn{2}{c}{Regular} & \multicolumn{2}{c}{DCNFIS}\\\cline{2-5}
		& AVG & STD & AVG & STD\\
		\hline
		CIFAR-10 & 96.620 & 0.23870 & \textbf{96.680} & 0.1789 \\
		\hline
		CIFAR-100 & \textbf{78.50}$^a$ & 0.1077 & 77.53 & 0.3523 \\
		\hline			
	\end{tabular}
\footnotesize{\\ \begin{flushleft} \hspace{1cm}	$a$-Significant at $\alpha$ = 0.05 \end{flushleft}}
\end{table}
}

\textcolor{black}{We compare the performance of DCNFIS with existing fuzzy methods in Table \ref{table_new}. This table is sorted based on accuracy of the methods on the CIFAR-10 dataset. As shown in this table, DCNFIS is presently the state-of-the-art neuro-fuzzy classifier on all four datasets.}

{
\addtolength{\tabcolsep}{-4pt}
\setlength\arrayrulewidth{0.25mm}
\setlength\extrarowheight{3pt}
\begin{table}[!t]
	\caption{\textcolor{black}{DCNFIS and Other Fuzzy Methods Accuracy Comparison }\label{table_new}}
	\centering
	{\color{black}\begin{tabular}{ccccc}
		\hline
		Method                   & MNIST          & Fashion-MNIST & CIFAR-10 $\downarrow$      & CIFAR-100      \\ \hline
		DCNFIS               & \textbf{99.59} & \textbf{94.4} & \textbf{96.68} & \textbf{77.53} \\ \hline
		Deep\_GK \cite{ref29,ref28} 	& 99.55 	& 92.03	& 91.87 \\ \hline
		\cite{riaz2019semi}       & 97.3           & -             & 88.2           & -              \\ \hline
		\cite{yazdanbakhsh2019deep}     & 99.58          & -             & 88.18          & 63.31          \\ \hline
		\cite{bodyanskiy2022deep}       & -              & 93.32         & 82.68          & -              \\ \hline
		\cite{diamantis2020fuzzy}     & 98.56          & 88.57         & 78.35          & -              \\ \hline
		\cite{beke2019interval}     & -              & -             & 77.49          & -              \\ \hline
		\cite{tan2023deep}     & 97.03          & 61.8          & 68.76          & 33.59          \\ \hline
		Deep\_FCM  \cite{ref30,ref28}           & 96.92          & 79.69         & 59.29          & -              \\ \hline
		\cite{shah2020adaptive}     & 96             & -             & 43.54          & -              \\ \hline
		\cite{di2021advanced}       & 98             & -             & 38             & -              \\ \hline
		\cite{sharma2019fuzzy}       & 93.44          & -             & 31.42          & -              \\ \hline
		GK   \cite{ref28}                & 81.61          & 73.1          & 28.59          & -              \\ \hline
		\cite{sharma2023mixed}       & 95.96          & -             & 27.83          & -              \\ \hline
		FCM \cite{ref30,ref28}     & 84.48          & 74.24         & 22.66          & -              \\ \hline
		\cite{zhang2019deep}       & -              & -             & 22.03          & -              \\ \hline
	\end{tabular}}
\end{table}
}

\section{\textcolor{black}{Evaluation on ImageNet}}

\subsection{\textcolor{black}{Dataset}}
\textcolor{black}{The ImageNet Large Scale Visual Recognition Challenge (ILSVRC) dataset \cite{russakovsky2015imagenet} is a subset of the ImageNet dataset \cite{JiaDeng2009}. The ImageNet dataset is organized according to the WordNet hierarchy \cite{miller1995wordnet}, and associates 21,841 synsets of WordNet with an average of 650 manually verified images representing those terms. As a result, ImageNet contains 14,197,122 annotated images organized by the semantic hierarchy of WordNet (as of August 2014). The ILSVRC subset includes 1,281,167 images for 1,000 distinct classes, with 732 to 1300 samples per class. ILSVRC also includes a labeled validation set of 50,000 images, which is usually used as the out-of-sample test set. (Note that, given the size of ILSVRC, cross-validation designs are not normally used for experiments with this dataset in the literature. For comparability, we also take the validation set as our out-of-sample test set.)}  

\subsection{\textcolor{black}{Experimental Setup}}
\textcolor{black}{In these experiments, we examine DCNFIS' performance with yet another CNN architecture, the Xception network \cite{chollet2017xception}. A pre-trained Xception network for ILSVRC is available, and researchers often begin transfer-learning projects using such a pretrained ILSVRC classifier. One of our experiments will thus focus on the performance of DCNFIS  with the Xception convolutional component when fine-tuning for ILSVRC; for a fair comparison, we will fine-tune the original Xception network as well, for the same number of epochs. In additon, we will conduct a new end-to-end training experiment on ILSVRC, again comparing the original Xception network with our DCNFIS.}

\textcolor{black}{Our DCNFIS\_Xception and regular Xception models are fine-tuned for 200 epochs using a batch size of 64, with weights initialized from the Xception model in Keras Applications \footnote{https://keras.io/api/applications/xception/}. Our learning rate is initially 0.001, with an annealing schedule that reduces it by a factor of 10 after 60, 100, 160, and 180 epochs. The regular Xception model is taken verbatim (excluding Dropout as discussed below) from Keras Applications; we freeze all layers in this model except the classifier, which is initialized with random weights. The DCNFIS version is also created by freezing the backend of the same pretrained model and replacing the classifier with our modified ANFIS, initialized with random weights.}

\textcolor{black}{Our end-to-end trained models, initialized with random weights throughout, will be first trained for 200 epochs with the same learning scheduler defined for the fine-tuned version. We apply a data augmentation pipeline created out of random rotation(0.05), random transition(0.1), random horizontal flip, and random zoom(-0.05, -0.15).} 

\subsection{\textcolor{black}{Evaluation of Results}}

\textcolor{black}{As shown in TABLE \ref{tab_ImageNet_ACC}, DCNFIS outperforms Xception in the fine-tuning experiment, with an out-of-sample accuracy of 77.62\% vs. 75.99\% in the base Xception model. In the end-to-end experiments, at 200 epochs the base Xception model outperforms DCNFIS, with an accuracy of 67.83 vs. 67.32 in the out-of-sample test. Xception had by this point achieved a training accuracy of over 99\%, while DCNFIS had only reached 87\%. However, the training accuracy for DCNFIS had not yet plateaued, and so it seemed reasonable to continue the experiment for additional epochs. We thus train the models for 200 more epochs with an initial learning rate of 1e-4, multiplied by 0.96 every 2 epochs. So in total, the models are trained for 400 epochs. We report the out-of-sample accuracies for Xception and DCNFIS at 300 and 400 epochs.} 

\textcolor{black}{The extended training of DCNFIS and Xception produces some very interesting results. As can be seen in TABLE \ref{tab_ImageNet_ACC}, the performance of Xception barely increases from 200 to 300 epochs of training, and not at all from 300 to 400 epochs. DCNFIS, on the other hand, improves by a large amount from 200 to 300 epochs, and a bit more from 300 to 400. At 400 epochs, DCNFIS is very substantially more accurate on the out-of-sample test than Xception, at 75.18\% accuracy vs. 67.92\%; a 10\% improvement on a test dataset of 50,000 images. We suspect the reason for these observations is the size of the DCNFIS classifier component; at 4 million parameters, it is itself quite substantial in size. It appears that this leads to improved modeling accuracy, at the cost of some increase in the number of training epochs needed. We thus claim that DCNFIS is more accurate than the base Xception network on ILSVRC. As far as we are aware, this is also the only application of \emph{fuzzy} classifiers to the ILSVRC dataset to date, and is thus the state-of-the-art in that respect.}  

\textcolor{black}{The reader will note that the accuracy results for Xception are lower than those reported in \cite{chollet2017xception}. In that paper, Dropout (with a 50\% dropout probability) is performed before the softmax layer in the classifier. As the Dropout technique is known to substantially improve generalization performance, and no Dropout method for our DCNFIS classifier currently exists, we omit Dropout in our fine-tuning and end-to-end training experiments reported above.}

{
\addtolength{\tabcolsep}{2pt}
\setlength\arrayrulewidth{0.25mm}
\setlength\extrarowheight{3pt}
	\begin{table}[!t]
		\caption{\textcolor{black}{Comparison of DCNFIS\_Xception with regular Xception on ImageNet} \label{tab_ImageNet_ACC}}
		\centering
		{\color{black}\begin{tabular}{cccccccc}
			\hline
			
%			\hline
			
			 \multicolumn{2}{c}{Training Type} & \multicolumn{2}{c}{Num. Epochs}&\multicolumn{2}{c}{Regular} & \multicolumn{2}{c}{DCNFIS} \\		
			
			\hline
			  \multicolumn{2}{c}{Fine-Tuning} & \multicolumn{2}{c}{200} & \multicolumn{2}{c}{75.99} & \multicolumn{2}{c}{\textbf{77.62}} \\
			
			\hline
			  \multicolumn{2}{c}{End-to-End} & \multicolumn{2}{c}{200} &\multicolumn{2}{c}{\textbf{67.83}} & \multicolumn{2}{c}{67.32} \\
			\hline

			  \multicolumn{2}{c}{End-to-End} & \multicolumn{2}{c}{300} & \multicolumn{2}{c}{67.92} & \multicolumn{2}{c}{\textbf{74.93}} \\
			\hline
			  \multicolumn{2}{c}{End-to-End} & \multicolumn{2}{c}{400} & \multicolumn{2}{c}{67.92} & \multicolumn{2}{c}{\textbf{75.18}} \\
			\hline			
		\end{tabular}}
	\end{table}
}

\section{Interpretability}

As discussed in section 4, DCNFIS uses the rule-based ANFIS architecture as its classifier component. Each rule in ANFIS is of the form of Eq. \ref{eq_8}, and forms a conjunction of antecedent clauses, each represented by one fuzzy subset of the corresponding input dimension. The rule itself thus defines a fuzzy region of the ANFIS input space. The collection of feature vectors within this fuzzy region can be considered a fuzzy cluster. Following \cite{ref28} we select the medoid element of a cluster as the representative for that cluster. The following equation shows the if-then rules of our modified ANFIS. \textcolor{black}{In this equation, $f_i(x)$ shows the $i^{th}$ feature in the extracted feature space for image $x$, and $N$ is the number of extracted features in the latent feature space, $M_{K,N}$ shows the $N^{th}$ membership for $K^{th}$ rule. The fuzzy region for each rule also is created by $M_{K,1}$ to $M_{K,N}$. Note that $\zeta_K$ is defined via eq. \ref{eq_4}.}

\textcolor{black}{
\begin{equation}
	\label{eq_8}
	\begin{array}{c}
		\displaystyle IF \ f_1(x)\ \ is\ M_{K,1}\ and\ f_2(x)\ is\ M_{K,2} \\ and \ \dots 
		\displaystyle  f_N(x) \ is\ \ M_{K,N}  \ \  THEN \\ \log{P\left(c_K\middle|x\right)} = \zeta_K - log(\sum_{i} \zeta_i) 		
		% where:\\
		% \zeta_K =  \omega_K +(\sum_j W_{jK} \cdot f_j(x) + b_K )
	\end{array}
\end{equation}
}
\begin{center}
	\begin{figure}[tb]
		\includegraphics[clip, trim=0.0cm 1.0cm 0.0cm 1.0cm, width=\linewidth]{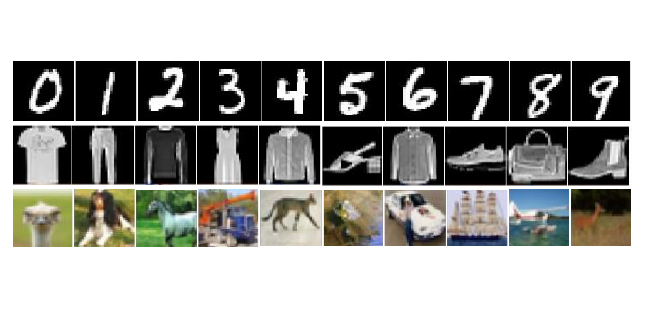}
		\caption{Class representatives derived from fuzzy rules. Top row: MNIST. Middle row: Fashion-MNIST. Bottom row: CIFAR-10}
		\label{fig_2}
	\end{figure}
\end{center}
In Fig. \ref{fig_2}, we present the medoid elements of the ten rules generated \textcolor{black}{for the MNIST, Fashion-MNIST, CIFAR-10 datasets}. Our approach to XAI is to treat the medoid as a synopsis of the entire cluster, \textcolor{black}{with the saliency map computed using \cite{Alber2018}\footnote{\href{https://github.com/albermax/innvestigate}{https://github.com/albermax/innvestigate}} for it constituting our explanans for the cluster. The general process is to train DCNFIS, and extract the fuzzy regions for each rule of the classifier. The region is treated as a fuzzy cluster, and the medoid element is identified. Next, we perform a saliency analysis on the medoids, using the Guided Backpropagation algorithm \cite{ref104}. As the medoid element is arguably the most representative of the whole class, we build our explanations on the saliency maps of the ten medoids. More details on the method, and an empirical evaluation of different saliency analyses, can be found in \cite{ref28,ref29}}. The power of DCNFIS in comparison with previous methods described in\cite{ref30,ref28,ref29} is that, by contrast, all the medoids are extracted from the rules of DCNFIS; there is no need for any further post training clustering and classification process.

We next demonstrate our medoid-based explanans on Fashion-MNIST. Our discussion in this section is inspired by our previous analysis of the MNIST Digits dataset in \cite{ref28}; we focus on examining selected misclassifications in the training dataset, comparing the saliency map of the erroneous examples against the saliency maps for the medoid images in the actual and predicted classes. As a contrast, we also examine selected correct classifications from the training dataset. Following this, we use UMAP visualizations to examine the misclassifications more formally. Finally, we discuss how our explanans were useful in detecting a learning bias being introduced in our model due to a commonly-used preprocessing step. 

Fig. \ref{fig_3} presents the class medoids derived from the fuzzy rules for the Fashion-MNIST dataset. The top row is the medoid element of each cluster/class (labeled from left to right T-shirt/Top, Trouser, Pullover, Dress, Coat, Sandal, Shirt, Sneaker, Bag, and Ankle Boot), and the bottom row is the corresponding saliency map.

\subsection{Fashion-MNIST}
\begin{figure}[tb]
	\vspace*{-1mm}\includegraphics[clip, trim=0.0cm 1.5cm 0.0cm 1.6cm, width=\linewidth]{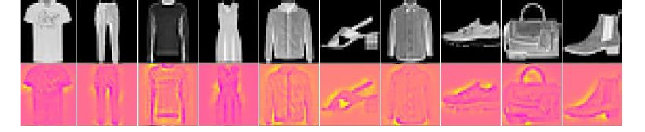}
	\caption{Class representatives derived from fuzzy rules for Fashion-MNIST dataset. First row: Medoids. Second row: Saliency of Medoids}
	\label{fig_3}
\end{figure}
\begin{figure}[tb]
	\centering
	\vspace*{-2mm}\includegraphics[clip, trim=0.5cm 0.5cm 1.0cm 0.5cm,width=0.5\linewidth]{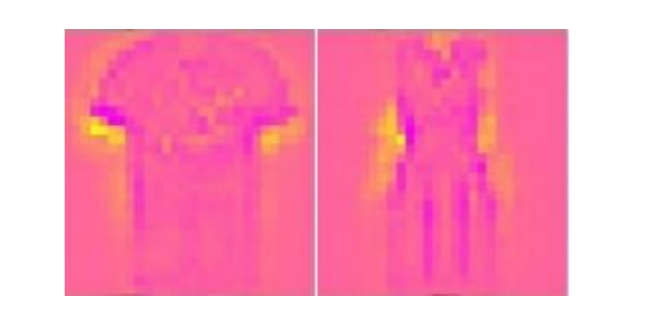}
	\caption{Difference between T-Shirt and Dress}
	\label{fig_4}
\end{figure}
One of the challenges with Fashion-MNIST is the similarity of some of the classes to each other. The images and labels were taken from the Zalando online store; the labels specifically are the “silhouette code” for that item, which is manually assigned by Zalando’s fashion staff (and then cross-checked by a separate team). Thus, while the labels are reliable, the distinctions between some classes are plainly finer than others. Coats and Pullovers, for instance, are only distinguished by the presence of a vertical zipper or line of buttons in the former, while the difference between a Coat and a Sandal is far more dramatic. The distinction between T-Shirt/Top and Dress is another example. As shown in Fig. \ref{fig_3}, the shapes formed by yellow pixels (strong negative impact on class assignment) are the main difference between these two classes. In Fig. \ref{fig_4} we have focused on this difference. For class Dress the neural network doesn’t care about having a short or long sleeve. It focuses on detection of a long almost vertical yellow line which starts from the axilla and ends at the high hip. For T-Shirts the vertical yellow pixels starting from the axilla peter out around the mid-torso. Meanwhile, a strong region of yellow pixels can be observed directly under the cutoff of each sleeve; long sleeves would thus strongly contraindicate the T-Shirt/Top class.

\begin{figure}[tb]
	\centering
	\includegraphics[clip, trim=0.5cm 0.7cm 1.5cm 0.4cm,width=\linewidth]{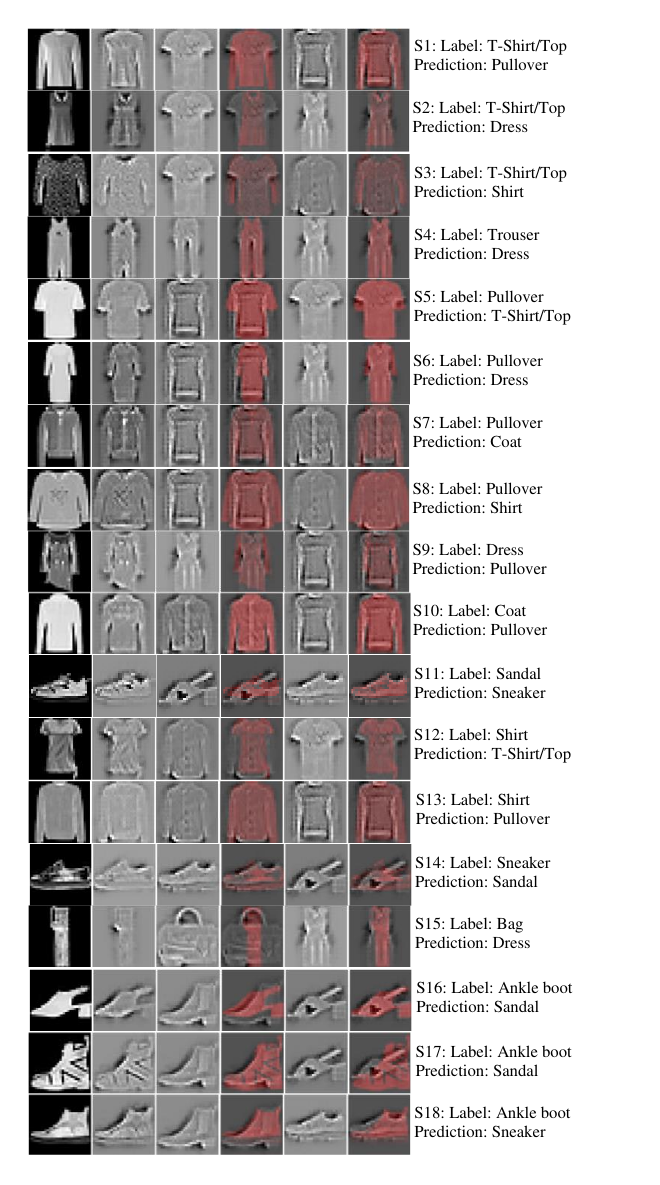}
	\caption{Images Misclassified by DCNFIS}
	\label{fig_5}
\end{figure}

Our analysis of DCNFIS misclassifications begins with an examination of misclassified images, presented in Fig. \ref{fig_5}. We selected 18 misclassified examples from the Fashion-MNIST training data. Note that this particular model has a training accuracy of 98.096 percent, so only 1,146 samples out of 60,000 are misclassified.  In this figure the original samples and their saliencies are shown in the first and second columns. In the third column we show the saliency of the medoid of the actual (label) class while in the fifth column we show the saliency of the predicted class. In columns four and six we overlay the saliencies of the sample on the saliencies from the label and prediction class medoids. For example, S1 is a T-Shirt/Top sample with long sleeves which is classified as a Pullover, and S2 is a T-Shirt/Top sample which has been found to be more similar to a Dress. This latter is an exemplar of a particularly challenging aspect of the T-Shirt/Top class, which we discuss in Section VI.B. What we see throughout these images is that the saliencies of each sample are noticeably more similar to the medoid of the predicted class (and the high-importance pixels especially so) than to the medoid of the labelled class. As in \cite{ref28}, this evidence tends to support our contention that the medoid saliences effectively capture the classification decisions of DCNFIS.

\begin{figure}[!t]
	\centering
	\includegraphics[clip, trim=0.4cm 0.2cm 0.4cm 1.4cm,width=\linewidth]{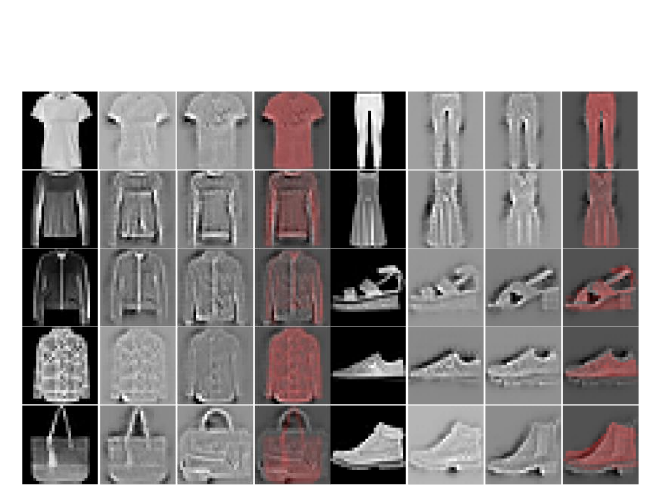}
	\caption{Samples Correctly Classified by DCNFIS}
	\label{fig_6}
\end{figure}

Fig. \ref{fig_6} shows 10 samples (one for each class) that are correctly classified by DCNFIS. Each row shows samples of two different classes. For each class, the first image shows the original sample, and the second image shows the saliency of the sample. The third image is the saliency of the class medoid, and the two saliences are overlaid in the fourth image, similar to columns of Fig. \ref{fig_5}. As shown in the figure, the sample and medoid saliencies are very similar. For example, an ankle boot (bottom right) sample has been classified by detection of the heel and upper surface of the boot.

Hendricks et al. \cite{ref105} proposed two criteria for explanations in image classification problems: they must be class discriminative, and image relevant. In that work, Hendricks et al. build natural language texts as explanations, so image relevance is a significant challenge. However, the more general point they raise is that explanations must refer to the specific content of the image in question. Understood in this fashion, we argue that our discussion above shows that the medoid-based approach is indeed image relevant. Indeed, each row of Fig. \ref{fig_5} compares the saliencies of a specific image against the actual and predicted class medoid saliencies, and highlights their similarities and differences. Class discriminativeness, meanwhile, is demonstrated in Fig. \ref{fig_3}.

\subsection{\textcolor{black}{Global explanation or local?}}
\begin{figure}[!t]
	\centering
	\subfloat{\includegraphics[scale=0.30,angle=90]{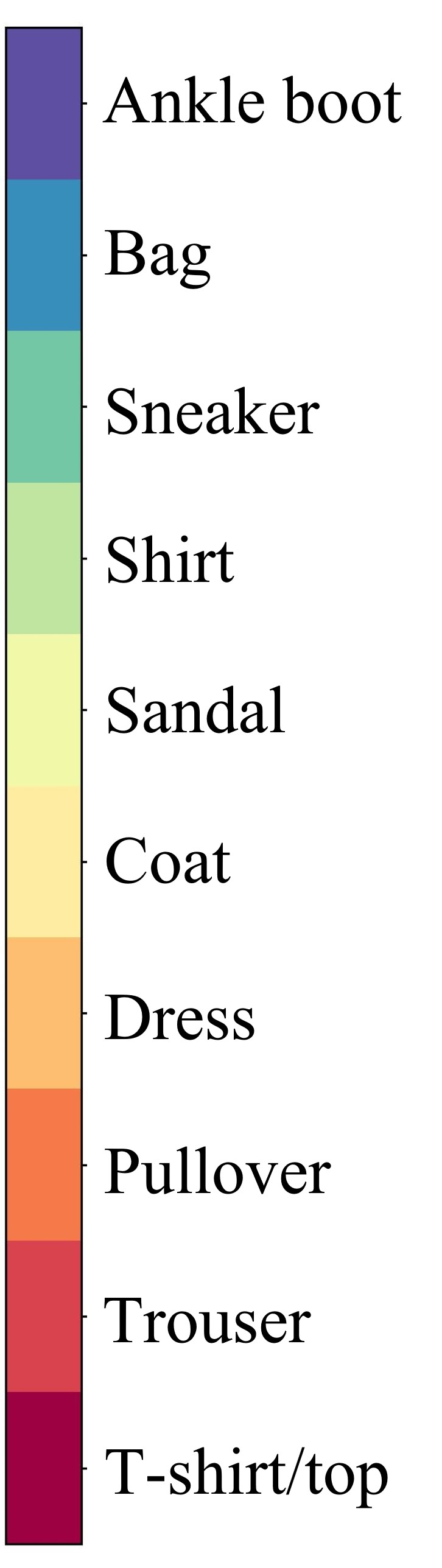}%
		\label{fig_7_3}} \\
	\subfloat{\vspace*{-5cm}\includegraphics[width=0.45\linewidth]{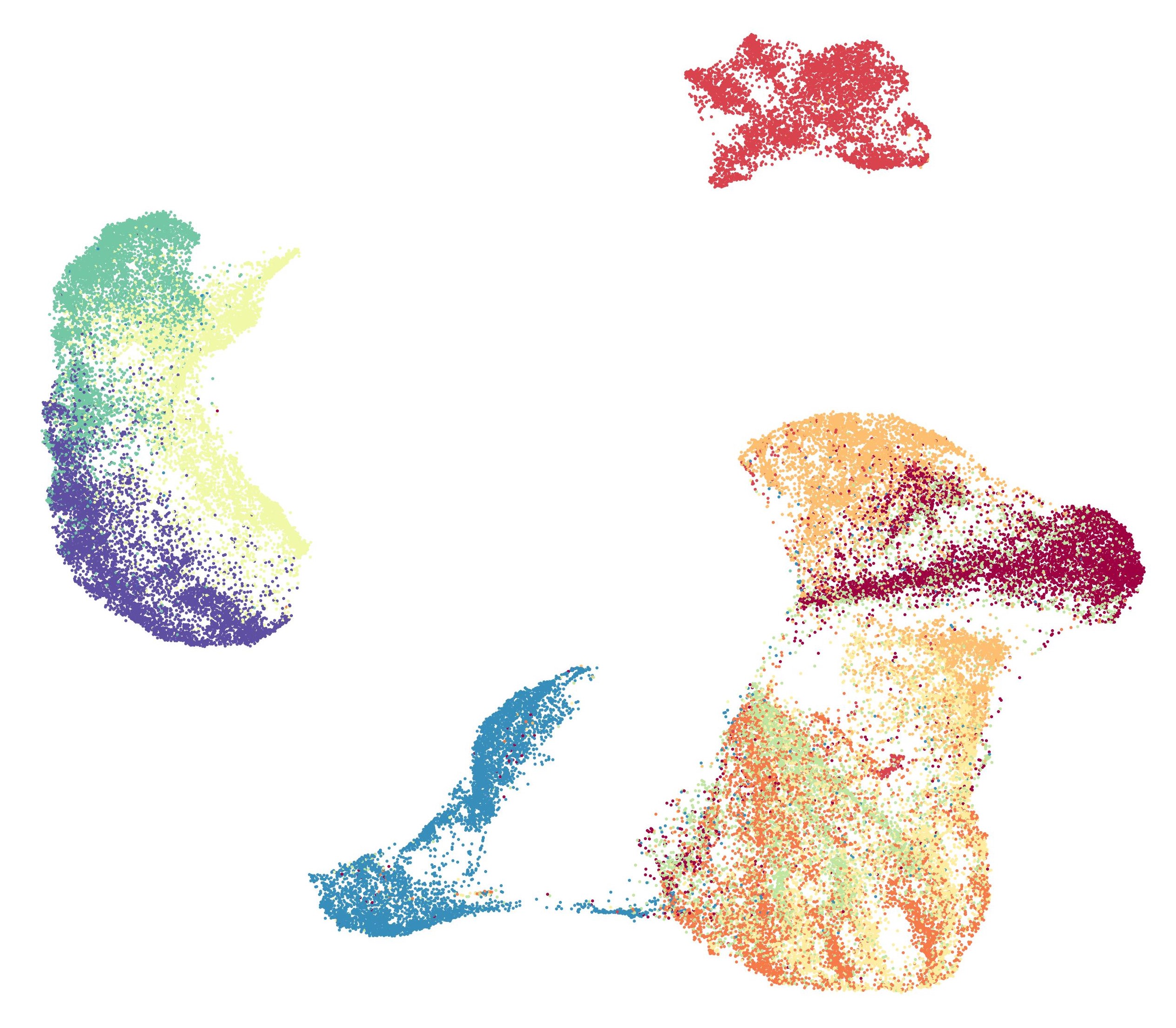}%
		\label{fig_7_1}}
	\subfloat{\includegraphics[scale=0.2,angle=0]{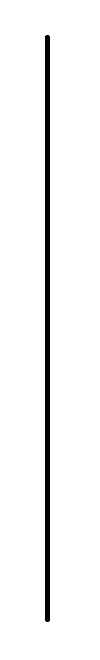}%
		\label{fig_7_4}}
	\subfloat{\includegraphics[width=0.5\linewidth]{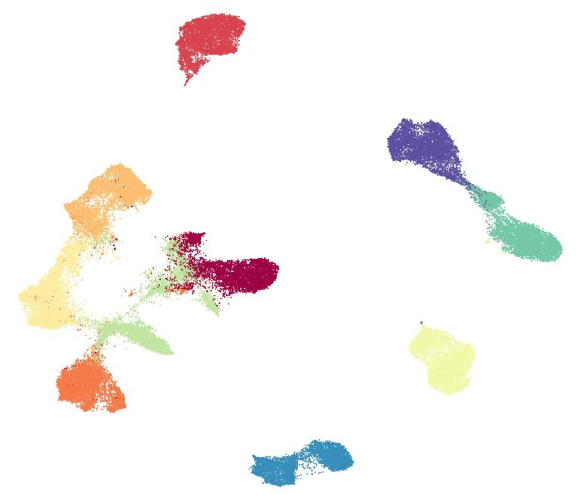}%
		\label{fig_7_2}}
	\caption{Left) 2-D UMAP visualization of Fashion-MNIST Dataset. Right) 2-D UMAP visualization of the DCNFIS Classifier Component Inputs (outputs from ResNet 29)}
	\label{fig_7}
\end{figure}

One of the principal advantages of CNNs is that the convolutional component acts as an automated feature selection algorithm. As DCNFIS does not change the convolution components of the base architecture, we expect that it will remain effective in this role. However, as DCNFIS is an end-to-end trainable algorithm, this expectation needs to be checked; the error signals being propagated through the fuzzy classifier component will, after all, be different than those propagated from layers of dense and SoftMax neurons. In this section, we thus evaluate this expectation on the Fashion-MNIST dataset, by visualizing the original dataset, and the outputs of the trained convolutional component, using the Uniform Manifold Approximation and Projection (UMAP) algorithm. We then further investigate selected misclassifications in these visualizations.

Left figure in Fig. \ref{fig_7} shows a 2-dimensional visualization of the Fashion-MNIST training dataset, created using UMAP \cite{ref106} with its default parameters. The Shirt, Coat, Dress, Pullover, and T-shirt/Top classes are badly comingled at the lower right, while the Ankle Boot, Sneaker and Sandal classes also overlap heavily at the upper left. Only the Trouser and Bag classes are well-separated.  Right figure in Fig. \ref{fig_7}  is a UMAP visualization (using the same parameters) of the training data output of the convolutional component of a trained DCNFIS network (i.e. the input to the fuzzy classifier). Plainly, there is a substantial improvement in the separation of the different classes (interpreting the compactness of the classes is more difficult, as UMAP is a nonlinear projection that balances preservation of local and global structure).

\begin{figure}[!t]
	\centering
	\vspace*{-2mm}\includegraphics[width=0.9\linewidth]{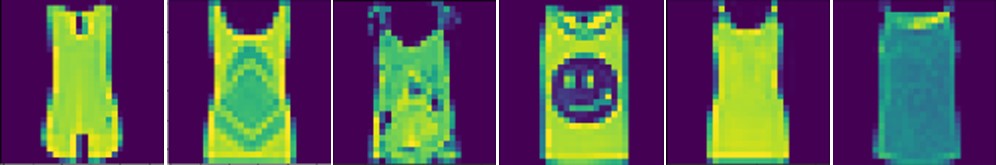}
	\vspace*{-1mm}\caption{Visualization of 6 samples of Top images out of 32}
	\label{fig_8}
\end{figure}
\begin{figure}[!t]
	\centering
	\vspace*{-2mm}\includegraphics[width=1.0\linewidth]{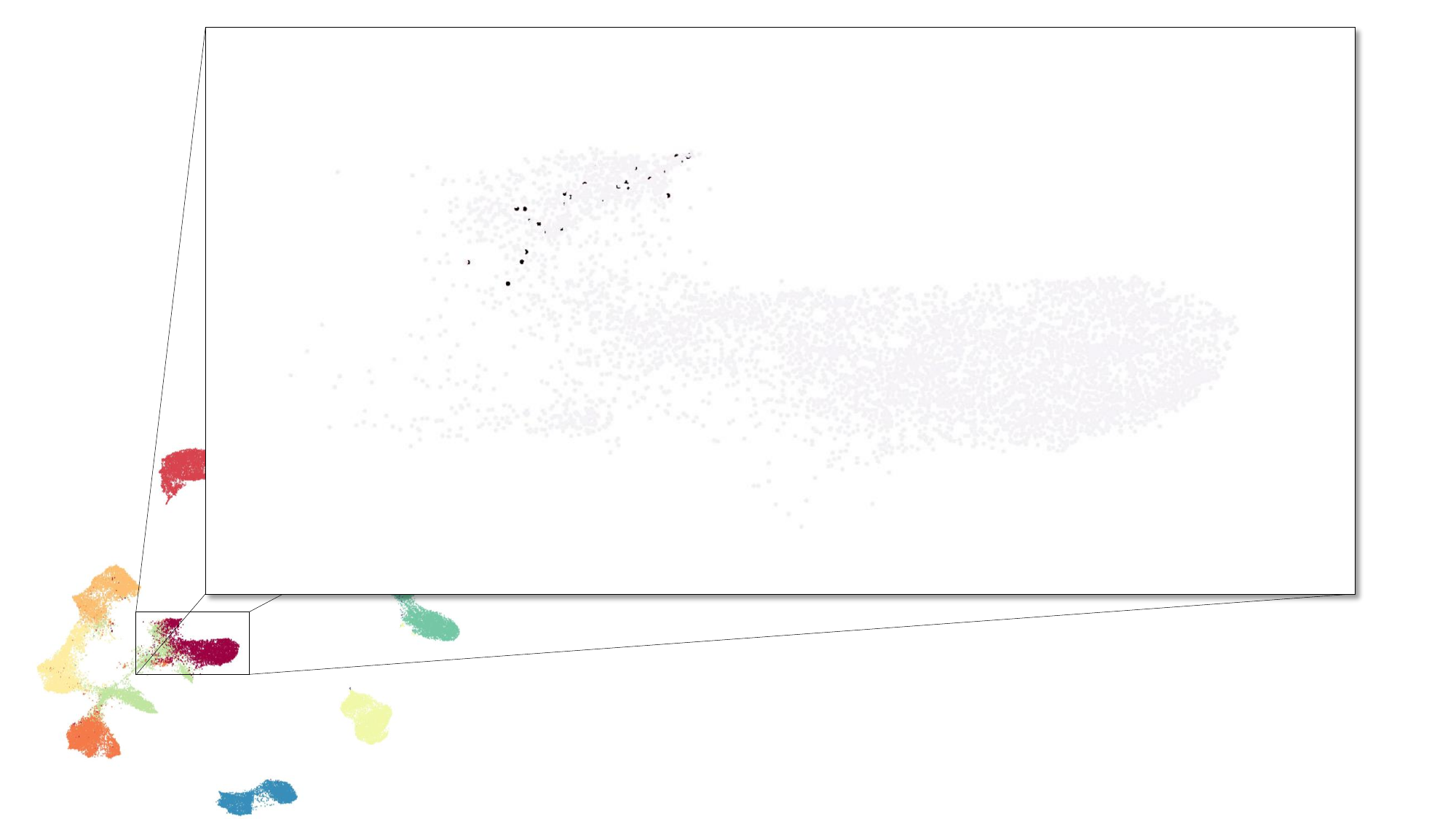}
	\vspace*{-4mm}\caption{Visualization of 32 selected Top samples out in the T-shirt/Top class region}
	\label{fig_9}
\end{figure}

However, an important characteristic of our medoid-based explanans can be observed in the T-shirt/Top class. We manually selected the sequentially first 32 images in this class that do not resemble T-shirts (i.e. that are sleeveless; Fig. \ref{fig_8}  presents 6 such examples). All but one of the 32 are correctly classified, but they do not closely resemble the T-shirt/Top medoid from Fig. \ref{fig_3} (leftmost image). When we highlight these 32 examples in Fig. \ref{fig_9}, we find that they all occur in the “upper” lobe of the class distribution, which appears to hold a concentration of examples at some distance from the class medoid. In this context, the misclassification of a Top as a Dress in Fig. \ref{fig_4} is revealing; sleeveless tops seem to lie in a different grouping from the class medoid. It is possible that splitting the T-shirt/Top class into separate clusters might yield a new medoid that better represents the Tops. Put another way, the T-shirt/Top class may well consist of multiple disjuncts. \textcolor{black}{In general, our medoid-based approach is likely to struggle with such classes, especially when the class is arguably a conflation of multiple real-world concepts. In these scenarios we should consider the explanation only as a local explanation as the medoid is not a good representative of the entire cluster.}

\subsection{Bias in Fashion-MNIST}
\begin{figure}[tb]
	\centering
	\includegraphics[clip, trim=0.0cm 0.0cm 0.0cm 0.0cm,width=0.8\linewidth]{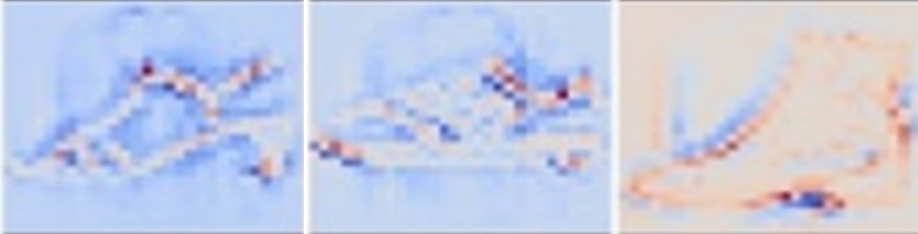}
	\caption{Medoids of DCNFIS for classes of Sandal, Sneaker, Ankle-Boot from training data with mean subtraction}
	\label{fig_10}
\end{figure}

One of the use cases for XAI is in debugging trained AI systems. Our experiments with DCNFIS unexpectedly offered a demonstration of this use case on the Fashion-MNIST dataset. One of the common preprocessing steps for image data is to subtract the per-pixel mean of the dataset from all images \cite{ref40}  (a zero-mean dataset is somewhat easier to train, as has been well-known for decades \cite{ref4}). However, in our experiments, we were finding that DCNFIS was noticeably less accurate than the base CNNs. When we look at our saliency maps, a reason for this appears, as in Fig. \ref{fig_10}. 

\begin{figure}[tb]
	\centering
	\includegraphics[clip, trim=0.0cm 0.0cm 0.0cm 0.0cm,width=0.3\linewidth]{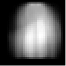}
	\caption{Mean of training data visualized as an image}
	\label{fig_11}
\end{figure}

What we see in these medoid saliencies (for the Sandal, Sneaker, and Ankle Boot classes) is what appears to be the faint image of some sort of top, indicated by pixels with negative saliency. However, no such structure appears in the actual images. The reason for this lies in an apparent bias of the Fashion-MNIST dataset; as has been noted, four of the ten classes are some sort of upper-body garment, and the Dress class is also similar. As a result, when we plot the per-pixel means as an image, we obtain the following (Fig. \ref{fig_11}).

{
\addtolength{\tabcolsep}{-4pt}
	\setlength\arrayrulewidth{0.25mm}
	\setlength\extrarowheight{3pt}
	\begin{table}[tb]
		\caption{Confusion Matrix of DCNFIS with mean subtraction\label{table2}}
		\centering
		\begin{tabular}{|c|c|c|c|c|c|c|c|c|c|c|}
			\hline
			& \rotatebox[origin=c]{270}{ \makecell{T-shirt\\top} } & \rotatebox[origin=c]{270}{Pullover} & \rotatebox[origin=c]{270}{Coat} & \rotatebox[origin=c]{270}{Shirt} & \rotatebox[origin=c]{270}{Dress} & \rotatebox[origin=c]{270}{Trouser} & \rotatebox[origin=c]{270}{Sandal} & \rotatebox[origin=c]{270}{Sneaker} & \rotatebox[origin=c]{270}{Ankle boot} & \rotatebox[origin=c]{270}{Bag} \\
			\hline
			\makecell{T-shirt\\top}&\cellcolor{green}5671& \cellcolor{cyan!80}41&2&\cellcolor{red}227&\cellcolor{cyan!80}56&0&1&0&0&2\\
			\hline
			Pullover& \cellcolor{cyan!80}58 & \cellcolor{green}5601 & \cellcolor{red}176 & \cellcolor{red}130 & \cellcolor{cyan!80}32 & 1 & 0 & 0 & 0 & 2\\
			\hline
			Coat& 1 & \cellcolor{cyan!80}29 &\cellcolor{green} 5833 & \cellcolor{red}84 & \cellcolor{cyan!80}49 & 2 & 0 & 0 & 0 & 2\\
			\hline
			Shirt& \cellcolor{red}326 & \cellcolor{cyan!80}94 & \cellcolor{cyan!80}109 & \cellcolor{green}5377 & \cellcolor{cyan!80}90 & 1 & 0 & 0 & 0 & 3\\
			\hline
			Dress&\cellcolor{cyan!80} 21 & 6 & \cellcolor{red}68 & \cellcolor{cyan!80}27 & \cellcolor{green}5875 & 2 & 0 & 0 & 0 & 1\\
			\hline
			Trouser& 1 & 1 & 1 & 3 & \cellcolor{red}15 & \cellcolor{green}5978 & 0 & 0 & 0 & 1\\
			\hline
			Sandal& 0 & 0 & 0 & 0 & 0 & 0 & \cellcolor{green}5981 & \cellcolor{red}16 & 0 & 2\\
			\hline
			Sneaker& 0 & 0 & 0 & 0 & 0 & 0 & \cellcolor{cyan!80}15 & \cellcolor{green}5925 & \cellcolor{red}60 & 0\\
			\hline
			\makecell{Ankle\\boot}& 0 & 0 & 0 & 0 & 0 & 0 & \cellcolor{cyan!80}19 & \cellcolor{red}105 &\cellcolor{green} 5874 & 2\\
			\hline
			Bag& 0 & 0 & 1 & 1 & 2 & 0 & 0 & 0 & 0 & \cellcolor{green}5996 \\
			\hline
		\end{tabular}
	\end{table}
}

{
\addtolength{\tabcolsep}{-4pt}
	\setlength\arrayrulewidth{0.25mm}
	\setlength\extrarowheight{3pt}
	\begin{table}[tb]
		\caption{Confusion Matrix of DCNFIS without mean subtraction\label{table3}}
		\centering
		\begin{tabular}{|c|c|c|c|c|c|c|c|c|c|c|}
			\hline
			& \rotatebox[origin=c]{270}{ \makecell{T-shirt\\top} } & \rotatebox[origin=c]{270}{Pullover} & \rotatebox[origin=c]{270}{Coat} & \rotatebox[origin=c]{270}{Shirt} & \rotatebox[origin=c]{270}{Dress} & \rotatebox[origin=c]{270}{Trouser} & \rotatebox[origin=c]{270}{Sandal} & \rotatebox[origin=c]{270}{Sneaker} & \rotatebox[origin=c]{270}{Ankle boot} & \rotatebox[origin=c]{270}{Bag} \\
			\hline
			\makecell{T-shirt\\top}&\cellcolor{green}5759&1\cellcolor{cyan!80}9&3&\cellcolor{red}192&\cellcolor{cyan!80}21&0&1&0&0&5\\
			\hline
			Pullover&\cellcolor{cyan!80}35&\cellcolor{green}5809&\cellcolor{red}75&\cellcolor{red}72&\cellcolor{cyan!80}9&0&0&0&0&0\\
			\hline
			Coat&0&\cellcolor{cyan!80}33&\cellcolor{green}5888&\cellcolor{red}55&\cellcolor{cyan!80}24&0&0&0&0&0\\
			\hline
			Shirt&\cellcolor{red}156&\cellcolor{cyan!80}75&\cellcolor{cyan!80}77&\cellcolor{green}5645&\cellcolor{cyan!80}44&1&0&0&0&2\\
			\hline
			Dress&\cellcolor{cyan!80}7&5&\cellcolor{red}28&\cellcolor{red}33&\cellcolor{green}5926&0&0&0&0&1\\
			\hline
			Trouser&0&0&0&1&4&\cellcolor{green}5995&0&0&0&0\\
			\hline
			Sandal&0&0&0&0&0&0&\cellcolor{green}5984&\cellcolor{red}13&3&0\\
			\hline
			Sneaker&0&0&0&0&0&0&\cellcolor{cyan!80}11&\cellcolor{green}5942&\cellcolor{red}46&1\\
			\hline
			\makecell{Ankle\\boot}&0&0&0&0&0&0&3&\cellcolor{red}84&\cellcolor{green}5913&0\\
			\hline
			Bag&0&0&1&1&1&0&0&0&0&\cellcolor{green}5997\\
			\hline
		\end{tabular}
	\end{table}
}

We see what seems like the silhouette of a long-sleeved top garment of some sort, with pant-like structures in the lower 2/3rds of the image. The outer edges of the pants, however, seem to line up with the presumptive axilla on the top garment, thus reinforcing its shape. This observation caused us to run a further set of experiments, without mean removal. In Tables \ref{table2} and TABLE \ref{table3}, we present the confusion matrices produced by DCNFIS on the training dataset with and without mean subtraction, respectively. Plainly, eliminating mean removal improved our accuracy; the total impact was four tenths of a percentage point on the test data, which is greater than the standard deviation of our ten replications (see Table \ref{table1}). In other words, the error introduced by mean subtraction was sufficient to significantly alter the outcomes of our experiments. Thus, our observations in this section serendipitously demonstrate that our medoid-based explanans on DCNFIS can be effective in debugging an AI

\section{Conclusion}
In this paper we have proposed a novel deep fuzzy network, which replaces the dense layers at the terminal end of a deep CNN with an ANFIS network. The architecture is end-to-end trainable, remains as accurate as the base CNNs is built from,\textcolor{black}{ and supports a medoid-based saliency-map explanation mechanism. In an exploratory study in the Fashion-MNIST dataset, the medoids derived from the fuzzy rules seem to form an effective global explanation for the DCNFIS model's classifications. This architecture can be applied in any of the domains the base CNNs are used in. We have also explored some novel applications of DCNFIS, in particular fault diagnosis of underground power cables (identifying the type and dgree of damage to a cable, not just whether damage has occured or not).}

\textcolor{black}{In future work, we will formalize and evaluate this explanation mechanism. A key step will be exploring splitting the rule clusters when multiple disjuncts are present; this may allow for more representative medoids to be extracted from the new clusters. We will also explore the use of triangular or trapezoidal memberships in DCNFIS; this may simplify the computation of a membership value, but would add discontinuities in the membership functions that would have to be incorporated in the error gradient calculations. Furthermore, we will examine how \emph{counterfactual} examples can be created from the DCNFIS rule clusters; counterfactuals are widely considered one of the most important classes of explanations for AI systems. When these extensions and refinements are complete, we believe that our EI will be ready for a comparative evaluation against other explanation mechanisms. In the XAI field, this necessarily takes the form of user studies comparing explanation "effectiveness" \cite{Gunning2021} amongst the different experimental contrasts.}

\textcolor{black}{An additional goal for our future work is to further improve the performance of DCNFIS, in particular by adding the Dropout algorithm. Rule Dropout was proposed in \cite{khalifa2016mcmi}; we believe that this can be combined with a standard Dropout technique in the linear consequent layer, resulting in superior generalization performance.}

\bibliographystyle{IEEEtran}

\bibliography{Bibliography}

\end{document}